\documentclass{bmvc2k}

\usepackage{graphicx}
\usepackage{amsmath,amssymb} 
\usepackage{amsfonts}       
\usepackage{xfrac}       
\usepackage{xcolor}
\usepackage{color}
\usepackage{mathtools}
\usepackage{commath}
\usepackage{nicefrac}
\usepackage{upgreek}

\def\eg{\emph{e.g}\bmvaOneDot}

\def\etal{\emph{et al}\bmvaOneDot}

\newcommand{\Cityscapes}{\textsc{Cityscapes}}
\newcommand{\Pascal}{\textsc{Pascal}}
\newcommand{\CoCo}{\textsc{COCO}}

\usepackage[normalem]{ulem}  
\def\clap#1{\hbox to 0pt{\hss #1\hss}}%
\newcommand{\para}[1]{\textbf{#1.}\hspace*{3mm}}
\newcommand{\bars}{|\hspace*{-0.25mm}|}
\newcommand{\bigbars}{\big|\hspace*{-0.375mm}\big|}
\newcommand{\odotx}{\hspace*{-0.35mm}\odot}

\title{Semi-supervised semantic segmentation needs strong, varied perturbations}

\addauthor{Geoff French}{g.french@uea.ac.uk}{1}
\addauthor{Samuli Laine}{slaine@nvidia.com}{2}
\addauthor{Timo Aila}{taila@nvidia.com}{2}
\addauthor{Michal Mackiewicz}{m.mackiewicz@uea.ac.uk}{1}
\addauthor{Graham aFinalyson}{g.finlayson@uea.ac.uk}{1}

\addinstitution{
School of Computing Sciences\\
University of East Anglia\\
Norwich, UK
}
\addinstitution{
NVIDIA\\
Helsinki, Finland
}

\runninghead{French et al.}{Semi-supervised semantic segmentation}

\begin{document}

\maketitle

\begin{abstract}
Consistency regularization describes a class of approaches that have yielded ground breaking results in semi-supervised classification problems. Prior work has established the cluster assumption\,---\,under which the data distribution consists of uniform class clusters of samples separated by low density regions\,---\,as important to its success. We analyze the problem of semantic segmentation and find that its' distribution does not exhibit low density regions separating classes and offer this as an explanation for why semi-supervised segmentation is a challenging problem, with only a few reports of success.
We then identify choice of augmentation as key to obtaining reliable performance without such low-density regions.
We find that adapted variants of the recently proposed CutOut and CutMix augmentation techniques
yield state-of-the-art semi-supervised semantic segmentation results in standard datasets.
Furthermore, given its challenging nature we propose that semantic segmentation acts as an effective \emph{acid test} for evaluating semi-supervised regularizers.
Implementation at: \url{https://github.com/Britefury/cutmix-semisup-seg}.
\end{abstract}

\section{Introduction}


Semi-supervised learning offers the tantalizing promise of training a machine learning model using datasets that have labels for only a fraction of their samples. These situations often arise in practical computer vision problems where large quantities of images are readily available and ground truth annotation acts as a bottleneck due to the cost and labour required. 

Consistency regularization \cite{Sajjadi:RegPertSemiSup,Laine:Temporal,Miyato:VATSemiSup,Oliver:RealisticEval} describes a class of semi-supervised learning algorithms that have yielded state-of-the-art results in semi-supervised classification, while being conceptually simple and often easy to implement. The key idea is to encourage the network to give consistent predictions for unlabeled inputs that are perturbed in various ways. 

The effectiveness of consistency regularization is often attributed to the \emph{smoothness assumption} \cite{Luo:SNTG} or \emph{cluster assumption} \cite{Chapelle:ClusterAssum,Sajjadi:Mutual,Shu:DIRTT,Verma:ICT}. The smoothness assumption states that samples close to each other are likely to have the same label. The cluster assumption\,---\,a special case of the smoothness assumption\,---\,states that decision surfaces should lie in low density regions of the data distribution. This typically holds in classification tasks, where most successes of consistency regularization have been reported so far.


At a high level, semantic segmentation is classification, where each pixel is classified based on its neighbourhood. It is therefore intriguing that there are only two reports of consistency regularization being successfully applied to segmentation from the medical imaging community \cite{Perone:SemiSupSeg,Li:SemiSupSkin} and none for natural photographic images.
We make the observation that the $L^2$ pixel content distance between patches centered on
neighbouring pixels varies smoothly even when the class of the center pixel changes, and thus there are no low-density regions along class boundaries.
This alarming observation leads us to investigate the conditions that can allow consistency regularization to operate in these circumstances. 

We find mask-based augmentation strategies to be effective for semi-supervised semantic segmentation,
with an adapted variant of CutMix \cite{Yun:CutMix} realizing significant gains.

The key contributions of our paper are our analysis of the data distribution of semantic segmentation and the simplicity of our approach.
We utilize tried and tested semi-supervised learning approaches,
and adapt CutMix -- an augmentation technique for supervised classification -- for semi-supervised learning and for segmentation,
achieving state of the art results.

\section{Background}
\label{sec:background}

Our work relates to prior art in three areas: recent regularization techniques for classification, semi-supervised classification with a focus on consistency regularization, and semantic segmentation.


\subsection{MixUp, Cutout, and CutMix}
\label{sec:background:mix}
The MixUp regularizer of Zhang \etal \cite{Zhang:MixUp} improves the performance of supervised image, speech and tabular data classifiers by using interpolated samples during training. The inputs and target labels of two randomly chosen examples are blended using a randomly chosen factor. 


The Cutout regularizer of Devries \etal \cite{Devries:Cutout} augments an image by masking a rectangular region to zero. The recently proposed CutMix regularizer of Yun \etal \cite{Yun:CutMix} combines aspects of MixUp and CutOut, cutting a rectangular region from image $B$ and pasting it over image $A$. MixUp, Cutout, and CutMix improve supervised classification performance, with CutMix outperforming the other two.

\subsection{Semi-supervised classification}
\label{sec:background:semisup}

A wide variety of consistency regularization based semi-supervised classification approaches have been proposed in the literature. They normally combine a standard supervised loss term (e.g. cross-entropy loss) with an unsupervised consistency loss term that encourages consistent predictions in response to perturbations applied to unsupervised samples.

The $\Pi$-model of Laine \etal \cite{Laine:Temporal} passes each unlabeled sample through a classifier twice, applying two realizations of a stochastic augmentation process, and minimizes the squared difference between the resulting class probability predictions.
%
%
Their temporal model and the model of Sajjadi \etal \cite{Sajjadi:RegPertSemiSup} encourage consistency between the current and historical predictions.
Miyato \etal \cite{Miyato:VATSemiSup} replaced the stochastic augmentation with adversarial directions, thus aiming perturbations toward the decision boundary.

The mean teacher model of Tarvainen \etal~\cite{Tarvainen:MeanTeachers} encourages consistency between predictions of a student network and a teacher network whose weights are an exponential moving average \cite{Polyak:Averaging} of those of the student. 
Mean teacher was used for domain adaptation in~\cite{French:SelfEnsDomAdapt}.



The Unsupervised data augmentation (UDA) model~\cite{Xie:UDA} and the state of the art FixMatch model~\cite{Sohn:FixMatch}
demonstrate the benefit of rich data augmentation as both combine CutOut \cite{Devries:Cutout} with
RandAugment \cite{Cubuk:RandAugment} (UDA) or CTAugment \cite{Berthelot:ReMixMatch} (FixMatch). RandAugment and CTAugment draw from a repertoire
of 14 image augmentations.

Interpolation consistency training (ICT) of Verma \etal \cite{Verma:ICT} and MixMatch \cite{Berthelot:MixMatch} both combine MixUp \cite{Zhang:MixUp} with consistency regularization. ICT uses the mean teacher model and applies MixUp to unsupervised samples, blending input images along with teacher class predictions to produce a blended input and target to train the student.


\subsection{Semantic segmentation}



Most semantic segmentation networks transform an image classifier into a fully convolutional network that produces a dense set of predictions for overlapping input windows, segmenting input images of arbitrary size~\cite{Long:FCN}. The DeepLab v3~\cite{Chen:DeepLabv3} architecture increases localization accuracy by combining atrous convolutions with spatial pyramid pooling. Encoder-decoder networks \cite{Badrinarayanan:SegNet,Ronneberger:UNet,Li:DenseUNet} use skip connections to connect an image classifier like encoder to a decoder. The encoder downsamples the input progressively, while the decoder upsamples, producing an output whose resolution natively matches the input.


A number of approaches for semi-supervised semantic segmentation use additional data. Kalluri \etal \cite{Kalluri:UnivSemiSupSeg} use data from two datasets from different domains, 
maximizing the similarity between per-class embeddings from each dataset. Stekovic \etal \cite{Stekovic:S4Net} use depth images and enforced geometric constraints between multiple views of a 3D scene.

Relatively few approaches operate in a strictly semi-supervised setting. Hung \etal \cite{Hung:AdvSemiSupSeg} and Mittal \etal \cite{Mittal:SSSHiLow} employ GAN-based adversarial learning,
using a discriminator network that distinguishes real from predicted segmentation maps to guide learning.

The only successful applications of consistency regularisation to segmentation that we are aware of come from the medical imaging community;
Perone \etal \cite{Perone:SemiSupSeg} and Li \etal \cite{Li:SemiSupSkin} apply consistency regularization to an MRI volume dataset and to skin lesions respectively. Both approaches use standard augmentation to provide perturbation.

\section{Consistency regularization for semantic segmentation}
\label{sec:cons_reg}

Consistency regularization adds a consistency loss term $L_\mathit{cons}$ to the loss that is minimized during training \cite{Oliver:RealisticEval}.
In a classification task, $L_\mathit{cons}$ measures a distance $d(\cdot, \cdot)$ between the predictions resulting from applying a neural network
$f_\theta$ to an unsupervised sample $x$ and a perturbed version $\hat{x}$ of the same sample, i.e., $L_\mathit{cons} = d(f_\theta(x), f_\theta(\hat{x}))$.
The perturbation used to generate $\hat{x}$ depends on the variant of consistency regularization used. A variety of distance measures
$d(\cdot, \cdot)$ have been used, e.g., squared distance \cite{Laine:Temporal} or cross-entropy \cite{Miyato:VATSemiSup}.


The benefit of the cluster assumption is supported by the formal analysis of Athiwaratkun \etal \cite{Athiwaratkun:ConsRegSWA}.
They analyze a simplified $\Pi$-model \cite{Laine:Temporal} that uses additive isotropic Gaussian noise for perturbation ($\hat{x} = x + \epsilon \mathcal{N}(0,1)$) and
find that the expected value of $L_\mathit{cons}$ is approximately proportional to the squared magnitude of the Jacobian $J_{f_\theta}(x)$ of the networks outputs with respect to its inputs.
Minimizing $L_\mathit{cons}$ therefore flattens the decision function in the vicinity of unsupervised samples, moving the 
decision boundary\,---\,and its surrounding region of high gradient\,---\,into regions of low sample density.



\subsection{Why semi-supervised semantic segmentation is challenging}
\label{sec:semseghard}

We view semantic segmentation as sliding window patch classification with the goal of identifying the class of the patch's central pixel.
Given that prior works~\cite{Laine:Temporal,Miyato:VATSemiSup,Sohn:FixMatch} apply perturbations to the raw pixel (input) space our analysis of the data distribution
focuses on the raw pixel content of image patches, rather than higher level features from within the network.

\begin{figure}[t]
\centering\footnotesize
\begin{tabular}{ccc}
\fbox{\includegraphics[width=0.30\textwidth]{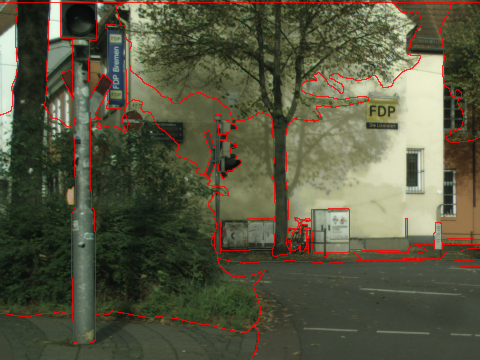}}&%
\fbox{\includegraphics[width=0.30\textwidth]{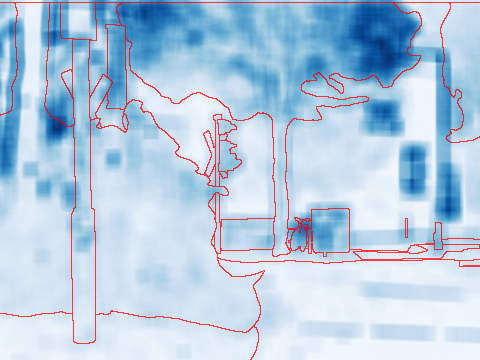}}&%
\fbox{\includegraphics[width=0.30\textwidth]{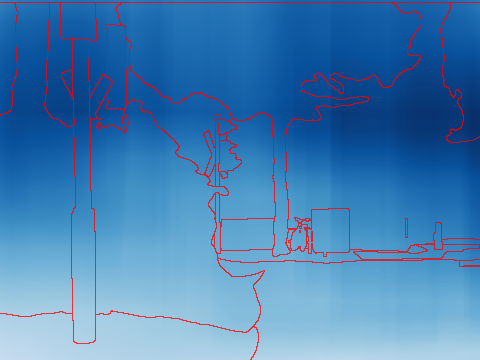}}\\
(a) Example image & (b) Avg. distance to neighbour, & (c) Avg. distance to neighbour,\\
& patch size 15$\times$15 & patch size 225$\times$225 \\
\end{tabular}
\caption{\label{fig:semseg:nolowdens:camvid_density}%
In a segmentation task, low-density regions rarely correspond to class boundaries.
(a) An image crop from the \Cityscapes{} dataset.
(b) Average $L^2$ distance between raw pixel contents of a patch centered at pixel $p$ and four overlapping patches centred on the immediate neighbours of $p$, using 15$\times$15 pixel patches.
(c) Same for a more realistic receptive field size of 225$\times$225 pixels.
A darker colour indicates larger inter-patch distance and therefore a low density region. Red lines indicate segmentation ground truth boundaries.
}
\end{figure}

We attribute the infrequent success of consistency regularization in natural image semantic segmentation problems 
to the observations that low density regions in input data do not align well with class boundaries.
The presence of such low density regions would manifest as locally larger than average $L^2$ distances between patches centred on neighbouring pixels that lie either side of a class boundary.
In Figure~\ref{fig:semseg:nolowdens:camvid_density} we visualise the $L^2$ distances between neighbouring patches.
When using a reasonable receptive field as in Figure~\ref{fig:semseg:nolowdens:camvid_density} (c) we can see that the cluster assumption is clearly violated:
how much the raw pixel content of the receptive field of one pixel differs from the contents of the receptive field of a neighbouring pixel has little correlation with whether the patches' center pixels belong to the same class.

The lack of variation in the patchwise distances is easy to explain from a signal processing perspective.
With patch of size $H \times W$, the distance map of $L^2$ distances between the pixel content of overlapping patches centered on all pairs of
horizontally neighbouring pixels can be written as \raisebox{0mm}[0mm][0mm]{$\sqrt{(\Delta_\mathrm{x}I)^{\circ 2} * 1^{H \times W}}$},
where $*$ denotes convolution and $\Delta_\mathrm{x}I$ is the horizontal gradient of the input image $I$.
The element-wise squared gradient image is thus low-pass filtered by a $H \times W$ box filter\footnote{We explain our derivation in our supplemental material}, which suppresses the fine details found in the high frequency
components of the image, leading to smoothly varying sample density across the image.

\begin{figure}[t]
\centering
\begin{tabular}{cc}
\includegraphics[width=0.5\textwidth]{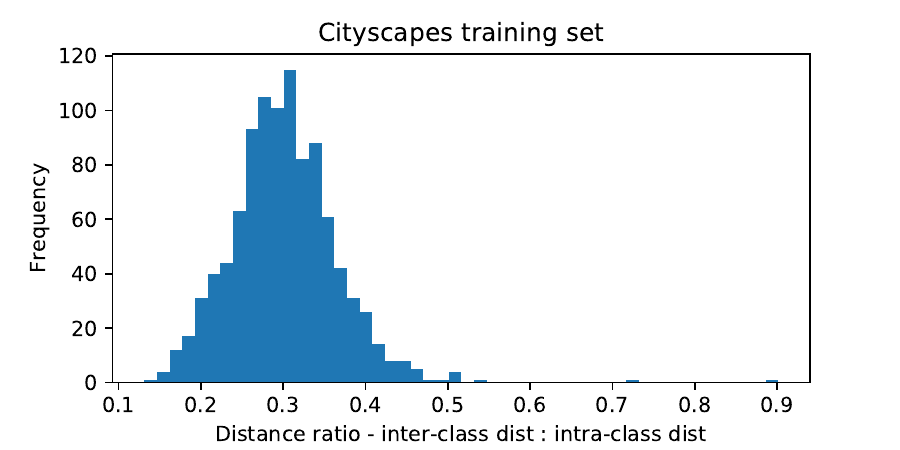} &
\includegraphics[width=0.4\textwidth,trim={0mm 160mm 291mm 0mm},clip]{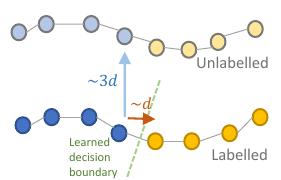}
\\
\end{tabular}
\caption{\label{fig:semseg:patch_cityscapes_dist_ratio}%
\textbf{Left:} histogram of the ratio $\nicefrac{|N_i - A_i|^2}{|P_i - A_i|^2}$ of the $L_2$ pixel content 
inter-class distance between patches $A_i$ and $N_i$ centred on neighbouring pixels either side of class boundary to the 
intra-class distance between nearest neighbour patches $A_i$ and $P_i$ coming from different images.
\textbf{Right:} conceptual illustration of semantic segmentation sample distribution.
The chain of samples (circles) below represents a row of patches from an image changing class (colour) half-way through.
The lighter chain above represents an unlabelled image. The dashed green line represents a learned decision boundary.
The samples within an image are at a distance of $\sim d$ from one another and $\sim 3d$ from those in another image.
}
\end{figure}

Our analysis of the \Cityscapes{} dataset quantifies the challenges involved in placing a decision boundary between two neighbouring pixels that should belong
to different classes, while generalizing to other images.
We find that the $L^2$ distance between patches centred on pixels on either side of a class boundary is $\sim \nicefrac{1}{3}$ of the distance to
the closest patch of the same class found in a different image (see Figure~\ref{fig:semseg:patch_cityscapes_dist_ratio}).
This suggests that precise positioning and orientation of the decision boundary are essential for good performance.
We discuss our analysis in further detail in our supplemental material.

\subsection{Consistency regularization without the cluster assumption}
\label{sec:cons_reg_noclus}

When considered in the context of our analysis above, the few reports of the successful application of consistency regularization
to semantic segmentation -- in particular the work of Li \etal~\cite{Li:SemiSupSkin} -- lead us to conclude that
the presence of low density regions separating classes is highly beneficial, but not essential.
We therefore suggest an alternative mechanism:
that of using non-isotropic natural perturbations such as image augmentation to
constrain the orientation of the decision boundary to lie parallel to the directions of perturbation (see the appendix
of Athiwaratkun \etal \cite{Athiwaratkun:ConsRegSWA}).
We will now explore this using a 2D toy example.

\begin{figure}[t]
\centering\footnotesize
\begin{tabular}{cc|cc}
\multicolumn{2}{c}{Isotropic perturbation}   &  \multicolumn{2}{|c}{\raisebox{0mm}[0mm][2mm]{Constrained perturbation}} \\
\includegraphics[width = 0.21\textwidth]{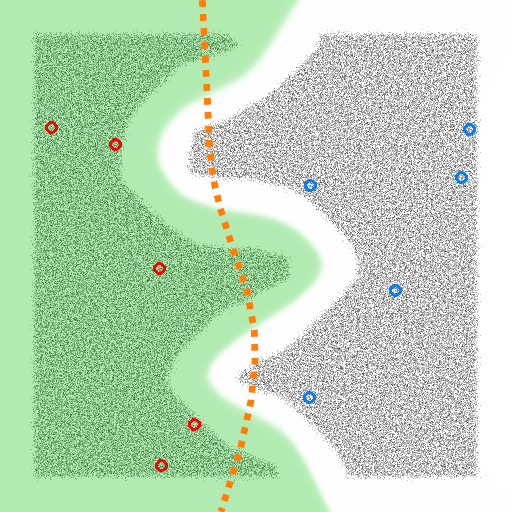} &
  \includegraphics[width = 0.21\textwidth]{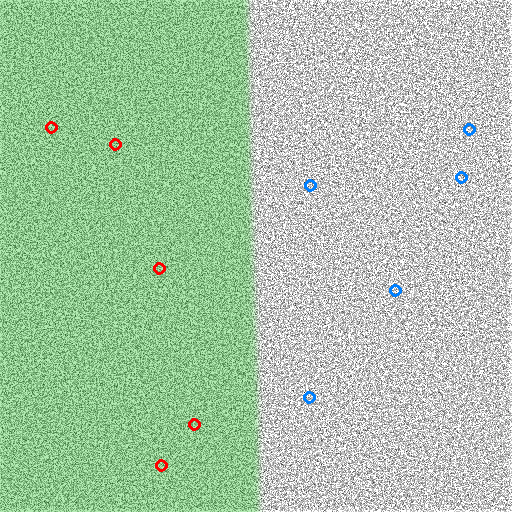} &
  \includegraphics[width = 0.21\textwidth]{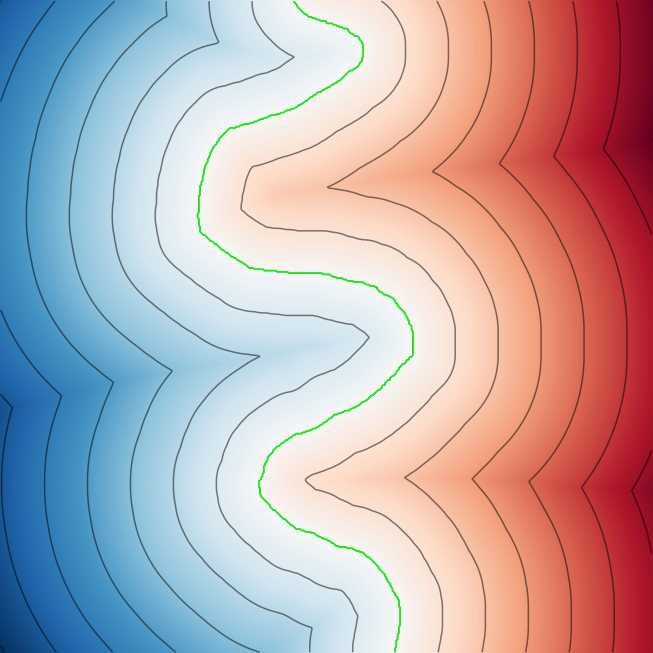} &
  \includegraphics[width = 0.21\textwidth]{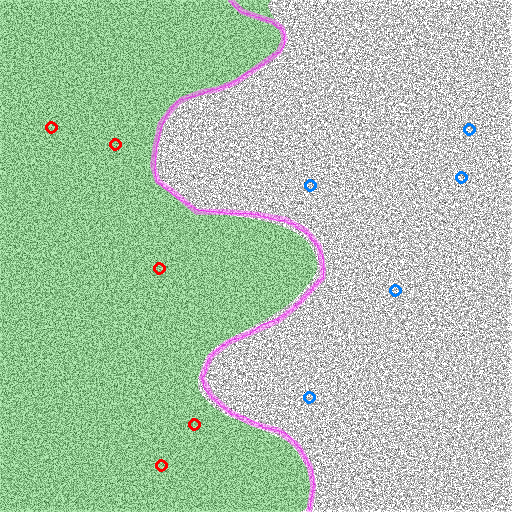} \\
(a) Low density region      & (b) No low density        & (c) Distance map   & (d) Constrain to dist.  \\
  separating classes     & region     &  and contours      &  map contours \\
\end{tabular}
\caption{\label{fig:semseg:clf2d}%
Toy 2D semi-supervised classification experiments. Blue and red circles indicate supervised samples from class 0 and 1 respectively. 
The field of small black dots indicate unsupervised samples. The learned decision function is visualized by rendering the probability
of class 1 in green.
(a, b) Semi-supervised learning with and without a low density region separating the classes. 
The dotted orange line in (a) shows the decision boundary obtained with plain supervised learning. 
(c) Rendering of the distance to the true class boundary with distance map contours.
Strong colours indicate greater distance to class boundary.
(d) Decision boundary learned when samples are perturbed along distance contours in (c). 
The magenta line indicates the true class boundary.
}
\end{figure}

Figure~\ref{fig:semseg:clf2d}a illustrates the benefit of the cluster assumption with a simple 2D toy mean teacher experiment, in which the cluster assumption holds due to the presence of a gap seperating the unsupervised samples that belong to two different classes.
The perturbation used for $L_\mathit{cons}$ is an isotropic Gaussian nudge to both coordinates, and as expected, the learned decision boundary settles neatly between the two clusters.
In Figure~\ref{fig:semseg:clf2d}b the unsupervised samples are uniformly distributed and the cluster assumption is violated.
In this case, the consistency loss does more harm than good; even
though it successfully flattens the neighbourhood of the decision function, it does so also across the
true class boundary. 


In Figure~\ref{fig:semseg:clf2d}c, we plot the contours of the distance to the true class boundary.
If we constrain the perturbation applied to a sample $x$ such that the perturbed $\hat{x}$ lies on or very close to the distance contour passing through $x$,
the resulting  learned decision boundary aligns well with the true class boundary, as seen in Figure~\ref{fig:semseg:clf2d}d.
When low density regions are not present the perturbations must be carefully chosen such that the probability of crossing
the class boundary is minimised.

We propose that reliable semi-supervised segmentation is achievable provided that the augmentation/perturbation mechanism 
observes the following guidelines: 1) the perturbations must be varied and high-dimensional in order to sufficiently
constrain the orientation of the decision boundary in the high-dimensional space of natural imagery, 2) the probability of a perturbation
crossing the true class boundary must be very small compared to the amount of exploration in other dimensions,
and 3) the perturbed inputs should be plausible; they should not be grossly outside the manifold of real inputs.

Classic augmentation based perturbations such as cropping, scaling, rotation and colour changes
have a low chance of confusing the output class and have proved to be effective in classifying natural
images~\cite{Laine:Temporal,Tarvainen:MeanTeachers}. Given that this approach has 
positive results in some medical image segmentation problems \cite{Perone:SemiSupSeg,Li:SemiSupSkin},
it is surprising that it is ineffective for natural imagery. This motivates us to search for stronger
and more varied augmentations for semi-supervised semantic segmentation.


\subsection{CutOut and CutMix for semantic segmentation}
\label{sec:cutout_cutmix}

Cutout~\cite{Devries:Cutout} yielded strong
results in semi-supervised classification in UDA \cite{Xie:UDA} and FixMatch \cite{Sohn:FixMatch}.
The UDA ablation study shows Cutout contributing the lions share of the semi-supervised performance, while
the FixMatch ablation shows that CutOut can match the effect of the combination of 14 image operations used
by CTAugment.
DeVries \etal~\cite{Devries:Cutout} established that Cutout encourages the network to utilise a wider variety
of features in order to overcome the varying combinations of parts of an image being present or masked out.
This variety introduced by Cutout suggests that it is a promising candidate for segmentation.

As stated in Section~\ref{sec:background:mix}, CutMix combines Cutout with MixUp, using a rectangular mask to blend input images. Given that
MixUp has been successfully used in semi-supervised classification in ICT \cite{Verma:ICT} and MixMatch
\cite{Berthelot:MixMatch}, we propose using CutMix to blend unsupervised samples and
corresponding predictions in a similar fashion.

Preliminary experiments comparing the $\Pi$-model \cite{Laine:Temporal} and the mean teacher model \cite{Tarvainen:MeanTeachers} indicate that
using mean teacher is essential for good performance in semantic segmentation, therefore all the experiments in this paper use the mean
teacher framework. We denote the student network as $f_\theta$ and the teacher network as $g_\phi$.




\para{Cutout}
As in \cite{Devries:Cutout} we initialize a mask $M$ with the value 1 and set the pixels inside a randomly chosen rectangle to 0.
To apply Cutout in a semantic segmentation task, we mask the input pixels with $M$ and disregard the consistency
loss for pixels masked to 0 by $M$. FixMatch \cite{Sohn:FixMatch} uses a \emph{weak} augmentation scheme consisting of crops and flips
to predict pseudo-labels used as targets for samples augmented using the \emph{strong} CTAugment scheme. Similarly,
we consider Cutout to be a form of
\emph{strong} augmentation, so we apply the teacher network $g_\phi$ to the original image to generate pseudo-targets that are
used to train the student $f_\theta$. Using square distance as the metric, we have
$L_\mathit{cons} = \bars{}M \odotx (f_\theta(M \odotx x) - g_\phi(x))\bars{}^2$, where $\odot$ denotes an
elementwise product.


\para{CutMix}
CutMix requires two input images that we shall denote $x_a$ and $x_b$ that we mix with the mask $M$. 
Following ICT (\cite{Verma:ICT}) we mix the teacher predictions for the input images $g_\phi(x_a), g_\phi(x_b)$ producing a pseudo target
for the student prediction of the mixed image. To simplify the notation, let us
define function $\mathit{mix}(a, b, M) = (1-M) \odotx a + M \odot b$ that selects the output pixel based on mask $M$.
We can now write the consistency loss as:
\begin{equation}
L_\mathit{cons} = \bigbars{}\mathit{mix}\big(g_\phi(x_a), g_\phi(x_b), M\big) - f_\theta\big(\mathit{mix}(x_a, x_b, M)\big)\bigbars{}^2\textrm{.}
\end{equation}

The original formulation of Cutout~\cite{Devries:Cutout} for classification used a rectangle of a fixed size and aspect ratio whose centre was positioned randomly,
allowing part of the rectangle to lie outside the bounds of the image. CutMix~\cite{Yun:CutMix} randomly varied the size, but used a fixed
aspect ratio. For segmentation we obtained better performance with CutOut by randomly choosing the size and aspect ratio and positioning the rectangle
so it lies entirely within the image. In contrast, CutMix performance was maximized by fixing the area of the rectangle to half that of the image, while
varying the aspect ratio and position.

While the augmentations applied by Cutout and CutMix do not appear in real-life imagery, they are reasonable from a visual standpoint.
Segmentation networks are frequently trained using image crops rather than full images, so blocking out a section of the image with Cutout can be seen as the inverse operation.
Applying CutMix in effect pastes a rectangular region from one image onto another, similarly resulting in a reasonable segmentation task.

Cutout and CutMix based consistency loss are illustrated in our supplemental material.


\section{Experiments}
\label{sec:experiments}

We will now describe our experiments and main results. We will start by describing the training setup, followed by results on the \Pascal{} VOC 2012, \Cityscapes{} and ISIC 2017 datasets.
We compare various perturbation methods in the context of semi-supervised semantic segmentation on \Pascal{} and ISIC.

\subsection{Training setup}
\label{sec:consseg:arch}

We use two segmentation networks in our experiments: 1) DeepLab v2 network \cite{Chen:DeepLabv2} based on ImageNet pre-trained ResNet-101 as used in \cite{Mittal:SSSHiLow} and 2) Dense U-net \cite{Li:DenseUNet} based on DensetNet-161 \cite{Huang:DenseNet} as used in \cite{Li:SemiSupSkin}. We also evaluate using DeepLab v3+ \cite{Chen:DeepLabv3plus} and PSPNet \cite{Zhao:PSPNet} in our supplemental material.

We use cross-entropy for the supervised loss $L_{sup}$ and compute the consistency loss $L_\mathit{cons}$ using the Mean teacher algorithm \cite{Tarvainen:MeanTeachers}.
Summing $L_\mathit{cons}$ over the class dimension and averaging over others allows us to minimize $L_{sup}$ and $L_\mathit{cons}$ with equal weighting. Further details and hyper-parameter settings are provided in supplemental material.
We replace the sigmoidal ramp-up that modulates $L_\mathit{cons}$ in \cite{Laine:Temporal, Tarvainen:MeanTeachers} with the average of the thresholded confidence of the teacher network, which increases as the training progresses \cite{French:SelfEnsDomAdapt, Sohn:FixMatch, Ke:DualStudent}.

\newcommand{\eb}[1]{\scriptsize\,$\pm$\,#1\normalsize}
\newcommand{\bp}{\makebox[0pt][l]{\textbf{\hspace*{-0.25mm}\%}}\phantom{\normalfont \%}}

\subsection{Results on Cityscapes and Augmented Pascal VOC}
\label{sec:consseg:res_cityscapes}

Here we present our results on two natural image datasets and contrast them against the state-of-the-art in semi-supervised semantic segmentation, which is currently the adversarial training approach of Mittal \etal\cite{Mittal:SSSHiLow}.  
We use two natural image datasets in our experiments. \Cityscapes{} consists of urban scenery and has 2975 images in its training set.
\Pascal{} VOC 2012\cite{Everingham:PascalVOC2012} is more varied, but includes only 1464 training images, and thus we follow the lead of Hung \etal\cite{Hung:AdvSemiSupSeg} and augment it
using \textsc{Semantic Boundaries}\cite{Hariharan:SemanticContours}, resulting in 10582 training images.
We adopted the same cropping and augmentation schemes as \cite{Mittal:SSSHiLow}.

In addition to an ImageNet pre-trained DeepLab v2, Hung \cite{Hung:AdvSemiSupSeg} and Mittal \etal\cite{Mittal:SSSHiLow}
also used a DeepLabv2 network pre-trained for semantic segmentation on the \CoCo{} dataset, whose natural image content is similar to that of \Pascal{}.
Their results confirm the benefits of task-specific pre-training.
Starting from a pre-trained ImageNet classifier is representative of practical problems for which a similar segmentation dataset is unavailable for pre-training,
so we opted to use these more challenging conditions only.

Our \Cityscapes{} results are presented in Table~\ref{tab:semseg:results:cityscapes} as mean intersection-over-union (mIoU) percentages, where higher is better.
Our supervised baseline results for \Cityscapes{} are similar to those of \cite{Mittal:SSSHiLow}. We attribute the small differences to training regime choices such as the choice of optimizer.
Both the Cutout and CutMix realize improvements over the supervised baseline, with CutMix taking the lead and improving on the adversarial\cite{Hung:AdvSemiSupSeg} and s4GAN\cite{Mittal:SSSHiLow} approaches.
We note that CutMix performance is slightly impaired when full size image crops 
are used getting an mIoU score of $58.75\%\pm0.75$ for 372 labelled images.
Using a mixing mask consisting of three smaller boxes -- see supplemental material -- whose scale better matches the image content alleviates this, obtaining $60.41\%\pm1.12$.

Our \Pascal{} results are presented in Table \ref{tab:semseg:results:pascalaug}.
Our baselines are considerably weaker than those of \cite{Mittal:SSSHiLow}; we acknowledge that we were unable to match them.
Cutout and CutMix yield improvements over our baseline and CutMix -- in spite of the weak baseline -- takes the lead, ahead of the adversarial and s4GAN results.
Virtual adversarial training~\cite{Miyato:VATSemiSup} yields a noticable improvement, but is unable to match competing approaches.
The improvement obtained from ICT~\cite{Verma:ICT} is just noticable, while standard augmentation makes barely any difference.
Please see our supplemental material for results using DeepLab v3+ \cite{Chen:DeepLabv3plus} and PSPNet \cite{Zhao:PSPNet} networks.

\newcommand{\RR}[1]{\raisebox{-0.25mm}{#1}}
\begin{table}[t]
\begin{center}%
\setlength{\tabcolsep}{2mm}%
\begin{tabular}{@{ }llllll@{ }}
\hline
\RR{Labeled samples}  			& \RR{\bf $\sim$1/30 (100)} & \RR{\bf 1/8 (372)}    & \RR{\bf 1/4 (744)}    & \RR{\bf All (2975)}  \\
\hline
\hline

&\multicolumn{4}{l}{\footnotesize{Results from \cite{Hung:AdvSemiSupSeg,Mittal:SSSHiLow} with ImageNet pretrained DeepLab v2}}     \\
Baseline                        & ---           			& 56.2\%				& 60.2\%                & 66.0\%               \\ 
Adversarial \cite{Hung:AdvSemiSupSeg} & ---           		& 57.1\%                & 60.5\%                & 66.2\%               \\ 
s4GAN \cite{Mittal:SSSHiLow}    & --- 						& 59.3\%                & 61.9\%                & 65.8\%               \\ 
\hline
&\multicolumn{4}{l}{\footnotesize{Our results: Same ImageNet pretrained DeepLab v2 network}}\\
Baseline              			& 44.41\%\eb{1.11}      	& 55.25\%\eb{0.66}      & 60.57\%\eb{1.13}      & 67.53\%\eb{0.35}     \\ 
Cutout                			& 47.21\%\eb{1.74}      	& 57.72\%\eb{0.83}      & 61.96\%\eb{0.99}      & 67.47\%\eb{0.68}     \\ 
CutMix                			& \bf51.20\%\eb{2.29}   	& \bf60.34\%\eb{1.24}   & \bf63.87\%\eb{0.71}   & \bf67.68\%\eb{0.37}  \\ 
\hline
\hline

\end{tabular}%
\caption{Performance (mIoU) on \Cityscapes{} validation set, presented as mean $\pm$ std-dev computed from 5 runs. The results for \cite{Hung:AdvSemiSupSeg} and \cite{Mittal:SSSHiLow} are taken from \cite{Mittal:SSSHiLow}.
}
\label{tab:semseg:results:cityscapes}
\end{center}
\end{table}

\begin{table}[t]
\begin{center}%
\setlength{\tabcolsep}{3mm}%
\begin{tabular}{@{ }llllll@{ }}
\hline
\RR{Labeled samples}   & \RR{\bf 1/100}        & \RR{\bf 1/50}         & \RR{\bf 1/20}         & \RR{\bf 1/8}          & \RR{\bf All (10582)}          \\ 
\hline
\hline
&\multicolumn{5}{l}{\footnotesize{Results from \cite{Hung:AdvSemiSupSeg,Mittal:SSSHiLow} with ImageNet pretrained DeepLab v2}}                  \\
Baseline                        & --           & 48.3\%                & 56.8\%               & 62.0\%                 & 70.7\%                \\ 
Adversarial \cite{Hung:AdvSemiSupSeg}  & --    & 49.2\%                & 59.1\%               & 64.3\%                 & 71.4\%                \\ 
s4GAN+MLMT \cite{Mittal:SSSHiLow}   & --       & 60.4\%                & 62.9\%               & 67.3\%                 & \bf73.2\%             \\ 
\hline
&\multicolumn{5}{l}{\footnotesize{Our results: Same ImageNet pretrained DeepLab v2 network}}\\
Baseline                        & 33.09\%      & 43.15\%               & 52.05\%              & 60.56\%                & 72.59\%               \\ 
Std. augmentation               & 32.40\%	   & 42.81\%   		  	   & 53.37\%              & 60.66\%                & 72.24\%               \\ 
VAT                             & 38.81\%      & 48.55\%               & 58.50\%              & 62.93\%                & 72.18\%               \\ 
ICT                             & 35.82\%      & 46.28\%               & 53.17\%              & 59.63\%                & 71.50\%               \\ 
CutOut                          & 48.73\%      & 58.26\%               & 64.37\%              & 66.79\%                & 72.03\%               \\ 
CutMix                          & \bf53.79\%   & \bf64.81\%            & \bf66.48\%           & \bf67.60\%             & 72.54\%               \\ 

\hline
\hline
\end{tabular}%
\caption{Performance (mIoU) on augmented \Pascal{}~VOC validation set, using same splits as Mittal \etal \cite{Mittal:SSSHiLow}. The results for \cite{Hung:AdvSemiSupSeg} and \cite{Mittal:SSSHiLow} are taken from \cite{Mittal:SSSHiLow}.
}
\label{tab:semseg:results:pascalaug}
\end{center}
\end{table}

%

\subsection{Results on ISIC 2017}
\label{sec:consseg:res_isic}

The ISIC skin lesion segmentation dataset~\cite{Codella:ISIC2017} consists of dermoscopy images focused on lesions set against skin. It has 2000 images in its training set and is a two-class (skin and lesion)
segmentation problem, featuring far less variation than \Cityscapes{} and \Pascal{}. 

We follow the pre-processing and augmentation schemes of Li \etal~\cite{Li:SemiSupSkin}; all images were scaled to $248\times248$ and our augmentation scheme consists of
random $224\times224$ crops, flips, rotations and uniform scaling in the range 0.9 to 1.1.

We present our results in Table~\ref{tab:semseg:results:isic}. We must first note that our supervised baseline results are noticably worse that those of Li \etal~\cite{Li:SemiSupSkin}.
Given this limitation, we use our results to contrast the effects of the different augmentation schemes used.
Our strongest semi-supervised result was obtained using CutMix, followed by standard augmentation, then VAT and CutOut.
We found CutMix to be the most reliable, as the other approaches required more hyper-parameter tuning effort to obtain positive resutlts.
We were unable to obtain reliable performance from ICT, hence its result is worse than that of the baseline.

We propose that the good performance of standard augmentation -- in contrast to \Pascal{} where it makes barely any difference -- is due to the lack of variation in the dataset.
An augmented variant of an unsupervised sample is sufficient similar to other samples in the dataset to successfully propagate labels, in spite of the limited varation introduced by standard augmentation.




\begin{table}[t]
\begin{center}%
\setlength{\tabcolsep}{3mm}%
\begin{tabular}{@{ }lllllll@{ }}
\hline
\RR{Baseline}      		& \RR{Std. aug.}			& \RR{VAT}			& \RR{ICT}			& \RR{Cutout}		& \RR{CutMix}       	& \RR{Fully sup.}       		\\ 
\hline
\hline
\multicolumn{6}{l}{\footnotesize{Results from \cite{Li:SemiSupSkin} with ImageNet pretrained DenseUNet-161}}                  					\\
72.85\%           		& \bf75.31\%              	& --              	& --              	& --              	& --                  	& 79.60\%					\\ 
\hline
\multicolumn{6}{l}{\footnotesize{Our results: ImageNet pretrained DenseUNet-161}}\\
67.64\%					& 71.40\%					& 69.09\%			& 65.45\%			& 68.76\%			& \bf74.57\%			& 78.61\%		\\
\eb{1.83}				& \eb{2.34}  				& \eb{1.38}			& \eb{3.50}  		& \eb{4.30}  		& \bf\eb{1.03}				& \eb{0.36}				\\

\hline

\hline
\end{tabular}%
\caption{Performance on ISIC 2017 skin lesion segmentation validation set, measured using the Jaccard index (IoU for lesion class). Presented as mean $\pm$ std-dev computed from 5 runs. All baseline and semi-supervised results
use 50 supervised samples. The fully supervised result ('Fully sup.') uses all 2000.
}
\label{tab:semseg:results:isic}
\end{center}
\end{table}

\subsection{Discussion}

We initially hypothesized that the strong performance of CutMix on the \Cityscapes{} and \Pascal{} datasets was due to the augmentation in effect `simulating occlusion',
exposing the network to a wider variety of occlusions, thereby improving performance on natural images.
This was our motivation for using the ISIC 2017 dataset; its' images do not feature occlusions and soft edges dilineate lesions
from skin\cite{Perez:LesionAug}. The strong performance of CutMix indicates that the presence of occlusions is not a requirement.

The success of virtual adversarial training demonstrates that exploring the space of adversarial examples provides sufficient variation to act as an effective semi-supervised
regularizer in the challenging conditions posed by semantic segmentation. In contrast the small improvements obtained from ICT and the barely noticable difference made by standard augmentation 
on the \Pascal{} dataset indicates that these approaches are not suitable for this domain; we recommend using a more varied source or perturbation, such as CutMix.

\section{Conclusions}
\label{sec:conc}

We have demonstrated that consistency regularization is a viable solution for semi-supervised semantic segmentation, provided that an appropriate source of augmentation is used.
Its data distribution lacks low-density regions between classes, hampering the effectiveness of augmentation schemes such as affine transformations and ICT.
We demonstrated that richer approaches can be successful, and presented an adapted CutMix regularizer that provides sufficiently varied perturbation to enable state-of-the-art results and work reliably on natural image datasets.
Our approach is considerably easier to implement and use than the previous methods based on GAN-style training.

We hypothesize that other problem domains that involve segmenting continuous signals given sliding-window input -- such as audio processing -- are likely to have similarly challenging distributions. This suggests
mask-based regularization as a potential avenue.

Finally, we propose that the challenging nature of the data distribution present in semantic segmentation indicates that it is an effective \emph{acid test} for evaluating future semi-supervised regularizers.

\section*{Acknowledgements}

Part of this work was done during an internship at nVidia. This work was in part funded under the European Union Horizon 2020 SMARTFISH project, grant agreement no. 773521.
Much of the computation required by this work was performed on the University of East Anglia HPC Cluster.
We would like to thank Jimmy Cross, Amjad Sayed and Leo Earl.
We would like thank nVidia coportation for their generous donation of a Titan X GPU.

%

\bibliography{cons_reg_bib}

%
%
%
%
%

\appendix

\section*{SUPPLEMENTAL MATERIAL}

\section{Pascal VOC 2012 performance across network architectures}

We demonstrate the effectiveness of our approach using a variety of architectures on the \Pascal{} dataset in Table~\ref{tab:sup:results:pascalaugarch}.
Using an ImageNet pre-trained DeepLab v3+ our baseline and semi-supervised results are stronger than those of \cite{Mittal:SSSHiLow}.

\begin{table}[h]
\begin{center}%
\setlength{\tabcolsep}{3mm}%
\begin{tabular}{@{ }llllll@{ }}
\hline
\RR{Prop. Labels}      & \RR{\bf 1/100}        & \RR{\bf 1/50}         & \RR{\bf 1/20}         & \RR{\bf 1/8}          & \RR{\bf Full (10582)}          \\ 
\hline
\hline
&\multicolumn{5}{l}{\footnotesize{Results from \cite{Hung:AdvSemiSupSeg,Mittal:SSSHiLow} with ImageNet pretrained DeepLab v2}}                  \\
Baseline                        & --           & 48.3\%                & 56.8\%               & 62.0\%                 & 70.7\%                \\ 
Adversarial \cite{Hung:AdvSemiSupSeg}  & --    & 49.2\%                & 59.1\%               & 64.3\%                 & 71.4\%                \\ 
s4GAN+MLMT \cite{Mittal:SSSHiLow}   & --       & 60.4\%                & 62.9\%               & 67.3\%                 & 73.2\%                \\ 
\hline
&\multicolumn{5}{l}{\footnotesize{Our results: Same ImageNet pretrained DeepLab v2 network}}\\
Baseline                        & 33.09\%      & 43.15\%               & 52.05\%              & 60.56\%                & 72.59\%               \\ 
CutMix                          & 53.79\%      & 64.81\%               & 66.48\%              & 67.60\%                & 72.54\%               \\ 

\hline
\hline
&\multicolumn{5}{l}{\footnotesize{Results from \cite{Mittal:SSSHiLow} with ImageNet pretrained DeepLab v3+}}                  \\
Baseline                        & --           & unstable              & unstable             & 63.5\%                 & 74.6\%                \\ 
s4GAN+MLMT \cite{Mittal:SSSHiLow}   & --       & 62.6\%                & 66.6\%               & 70.4\%                 & 74.7\%                \\ 
\hline
&\multicolumn{5}{l}{\footnotesize{Our results: ImageNet pretrained DeepLab v3+ network}}\\
Baseline                        & 37.95\%      & 48.35\%               & 59.19\%              & 66.58\%                & 76.70\%               \\ 
CutMix                          & 59.52\%      & 67.05\%               & 69.57\%              & 72.45\%                & 76.73\%               \\ 

\hline
\hline
&\multicolumn{5}{l}{\footnotesize{Our results: ImageNet pretrained DenseNet-161 based Dense U-net}}\\
Baseline                        & 29.22\%      & 39.92\%               & 50.31\%              & 60.65\%                & 72.30\%               \\ 
CutMix                          & 54.19\%      & 63.81\%               & 66.57\%              & 66.78\%                & 72.02\%               \\ 

\hline
\hline
&\multicolumn{5}{l}{\footnotesize{Our results: ImageNet pretrained ResNet-101 based PSPNet}}\\
Baseline                        & 36.69\%      & 46.96\%               & 59.02\%              & 66.67\%                & 77.59\%               \\ 
CutMix                          & 67.20\%      & 68.80\%               & 73.33\%              & 74.11\%                & 77.42\%               \\ 

\hline
\end{tabular}%
\caption{Performance (mIoU) on augmented \Pascal{}~VOC validation set across a variety of architectures, using same splits as Mittal \etal \cite{Mittal:SSSHiLow}.
The results for \cite{Hung:AdvSemiSupSeg} and \cite{Mittal:SSSHiLow} are taken from \cite{Mittal:SSSHiLow}.
}
\label{tab:sup:results:pascalaugarch}
\end{center}
\end{table}

\section{Smoothly varying sample density in semantic segmentation}
\label{sec:supp:smooth}

\subsection{Derivation of signal processing explanation}

\begin{figure}[h]
\centering
\begin{tabular}{ccc}
(a) Patch $A$ & (b) Patch $B$ & (c) Patch from $\Delta_\mathrm{x}I$ \\
\includegraphics[width=0.3\textwidth]{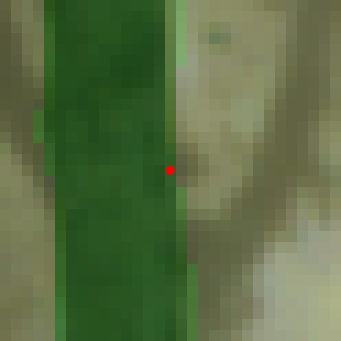} &
\includegraphics[width=0.3\textwidth]{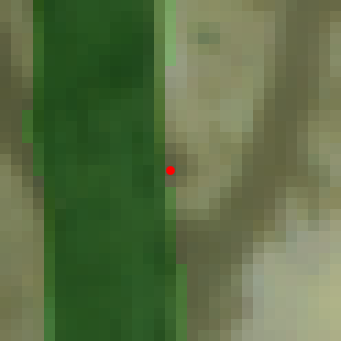} &
\includegraphics[width=0.3\textwidth]{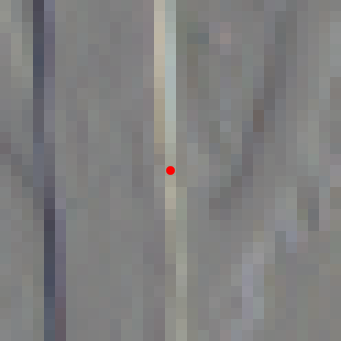} \\
\end{tabular}
\caption{\label{fig:semseg:patch_cityscapes_ab}
(a, b) Two patches centred on horizontally neighbouring pixels, extracted from the Cityscapes Image in Figure~\ref{fig:semseg:nolowdens:camvid_density}(a).
The ground truth vegetation class is overlayed in green. The red dot indicates the central pixel.
(c) The same patch extracted from the horizontal gradient image.
}
\end{figure}

In this section we explain the derivation of our signal-processing based explanation of the lack of low-density regions in semantic segmentation problems.

To analyse the smoothness of the distribution of patches over an image we need to compute the $L^2$ pixel content distance between patches centred on neighbouring pixels.
Let us start with two patches $A$ and $B$ -- see Figure~\ref{fig:semseg:patch_cityscapes_ab}(a,b) -- extracted from an image $I$, centred on horizontally neighbouring pixels, with $A$ one pixel to the left of $B$.
The $L^2$ distance is $|B - A|$.
Given that each pixel in $B - A$ is the difference between horizontally neighbouring pixels, $B - A$ is therefore a patch extracted from the horizontal gradient
image $\Delta_\mathrm{x}I$ (see Figure~\ref{fig:semseg:patch_cityscapes_ab}(c)).
The squared distance is the sum of the element-wise squares of $B - A$; it is the sum of the elements in a patch extracted from $(\Delta_\mathrm{x}I)^{\circ 2}$.
Computing the sums of all patches of size $H\times W$ in a sliding window fashion across $(\Delta_\mathrm{x}I)^{\circ 2}$ is equivalent to convolving it with a box kernel $1^{H \times W}$, thus the
distance between all horizontally neighbouring patches can be computed using \raisebox{0mm}[0mm][0mm]{$\sqrt{(\Delta_\mathrm{x}I)^{\circ 2} * 1^{H \times W}}$}.
A box filter -- or closely related uniform filter -- is a low-pass filter that will suppress high-frequency details, resulting in a smooth output.
This is implemented in a Jupyter notebook~\cite{Kluyver:JupyterNotebook} that is distributed with our code.

\subsection{Analysis of patch-to-patch distances within Cityscapes}

Our analysis of the \Cityscapes{} indicates that semantic segmentation problems exhibit \emph{high intra}-class variance and \emph{low inter}-class variance.
We chose 1000 image patch triplets each consisting of an anchor patch $A_i$ and positive $P_i$ and negative $N_i$ patches with the same and different ground truth classes as $A_i$ respectively.
We used the $L^2$ pixel content intra-class distance $|P_i - A_i|^2$ and inter-class distance $|N_i - A_i|^2$ as proxies for variance.
Given that a segmentation model must place a decision boundary between neighbouring pixels of different classes within an image we chose $A_i$ and $N_i$ to be immediate neighbours on either side of a class boundary.
As the model must also generalise from a labelled images to unlabelled images we searched all images except that containing $A_i$ for the $P_i$ belonging to the same class that minimises $|P_i - A_i|^2$. Minimising the distance
chooses the best case intra-class distance over which the model must generalise.
The inter-class to intra-class distance ratio histogram on the left of Figure~\ref{fig:semseg:patch_cityscapes_dist_ratio} underlies the illustration to the right in which the blue intra-class distance is approximately $3\times$
that of the red inter-class distance. The model must learn to place the decision boundary between the patches centred on neighbouring pixels, while orienting it sufficiently accurately that it intersects other images at the correct
points.

\section{Setup: 2D toy experimnents}

The neural networks used in our 2D toy experiments are simple classifiers in which samples are 2D $x,y$ points ranging from -1 to 1. Our networks are multi-layer perceptrons consisting of 3 hidden layers of 512 units, each followed by a ReLU non-linearity. The final layer is a 2-unit classification layer. We use the mean teacher \citep{Tarvainen:MeanTeachers} semi-supervised learning algorithm with binary cross-entropy as the consistency loss function, a consistency loss weight of 10 and confidence thresholding \citep{French:SelfEnsDomAdapt} with a threshold of 0.97.

The ground truth decision boundary was derived from a hand-drawn 512$\times$512 pixel image. The distance map shown in Figure~\ref{fig:semseg:clf2d}(c) was computed using the \path{scipy.ndimage.morphology.distance_transform_edt} function from SciPy~\cite{SciPy}, with distances negated for regions assigned to class 0. Each pixel in the distance map therefore has a signed distance to the ground truth class boundary. This distance map was used to generate the countours seen as lines in Figure~\ref{fig:semseg:clf2d}(c) and used to support the constrained consistency regularization experiment illustrated in Figure~\ref{fig:semseg:clf2d}(d).

The constrained consistency regularization experiment described in Section~\ref{sec:cons_reg_noclus} required that a sample $x$ should be perturbed to $\hat{x}$ such that they are at the same\,---\,or similar\,---\,distance to the ground truth decision boundary. This was achieved by drawing isotropic perturbations from a normal distrubtion $\hat{x} = x + h$ where $h \sim \mathcal{N}(0, 0.117)$ ($0.117 \approx 30$ pixels in the source image), determining the distances $m(x)$ and $m(\hat{x})$ from $x$ and $\hat{x}$ to the ground truth boundary (using a pre-computed distance map) and discarding the perturbation -- by masking consistency loss for $x$ to 0 -- if $|m(\hat{x}) - m(x)| > 0.016$ ($0.016 \approx 4$ pixels in the source image).

\section{Semantic segmentation experiment setup}

\subsection{Adapting semi-supervised classification algorithms for segmentation}
\label{sec:supp:seg:adapt}

In the main paper we explain how we adapted Cutout~\cite{Devries:Cutout} and CutMix~\cite{Yun:CutMix} for segmentation.
Here we will discuss our approach to adapting standard augmentation, Interpolation Consistency Training (ICT) and Virtual Adversarial Training (VAT).
We note that implementations of all of these approaches are supplied with our source code.

\subsubsection{Standard augmentation}

Our standard augmentation based consistency loss uses affine transformations to modify unsupervised images.
Applying different affine transformations within the teacher and student paths results in predictions that not aligned.
An appropriate affine transformation must be used to bring them into alignment.
To this end, we follow the approch used by Perone \etal~\cite{Perone:SemiSupSeg} and Li \etal~\cite{Li:SemiSupSkin};
the original unaugmented image $x$ is passed to the teacher network $g_\phi$ producting predictions $g_\phi(x)$, aligned with
the original image. The image is augmented with an affine transformation $a(\cdot)$: $\hat{x} = a(x)$, which is passed to the student network $f_\theta$
producting predictions $f_\theta(a(x))$. The same transformation is applied to the teacher prediction: $a(g_\phi(x))$. The two predictions are
now geometrically aligned, allowing consistency loss to be computed.

At this point we would like to note some of the challenges involved in the implementation.
A natural approach would be to use a single system for applying affine transformations, e.g. the affine grid functionality provided by PyTorch~\cite{PyTorch};
that way both the input images and the predictions can be augmented using the same transformation matrices.
We however wishe to exactly match the augmentation system used by Hung \etal~\cite{Hung:AdvSemiSupSeg} and Mittal \etal~\cite{Mittal:SSSHiLow}, both
of which use functions provided by OpenCV~\cite{OpenCV}. This required gathering a precise understanding of how the relevant functions in
OpenCV generate and apply affine transformation matrices in order to match them using PyTorch's affine grid functionality, that must be used to
transform predictions.

\subsubsection{Interpolation consistency training}

ICT was the simplest approach to adapt. We follow the procedure in \cite{Verma:ICT}, except that our networks generate pixel-wise class probability vectors.
These are blended and loss is computed from them in the same fashion as \cite{Verma:ICT}; the only different is that the arrays/tensors have additional dimensions.

\subsubsection{Virtual Adversarial Training}

Following the notation of Oliver \etal~\cite{Oliver:RealisticEval}, in a classification scenario VAT computes the adversarial perturbation $r_{adv}$ as:

$$r \sim \mathcal{N}(0, \frac{\upxi}{\sqrt{dim(x)}}I)$$

$$g = \nabla_r d(f_\theta(x), f_\theta(x + r))$$

$$r_{adv} = \epsilon \frac{g}{||g||}$$

We adopt exactly the same approach, computing the adversarial perturbation that maximises the mean of the change in class prediction for all pixels of the output.

We scale the adversarial radius $\epsilon$ adaptively on a per-image basis by multiplying it by the magnitude of the gradient of the input image.
We find that a scale of 1 works well and used this in our experiments.
We also tried using a fixed value for $\epsilon$ -- as normally used in VAT -- and found that doing so caused a slight but statistically insignificant reduction in performance.
We therefore recommend the adaptive radius on the basis of ease of use. It is implemented in our source code.

\subsection{Illustration of computation of CutMix and Cutout}
\label{app:seg:cutmix_illus}

\begin{figure}[t]
\centering
\includegraphics[width=.9\textwidth,trim={2mm 91mm 129mm 3.5mm},clip]{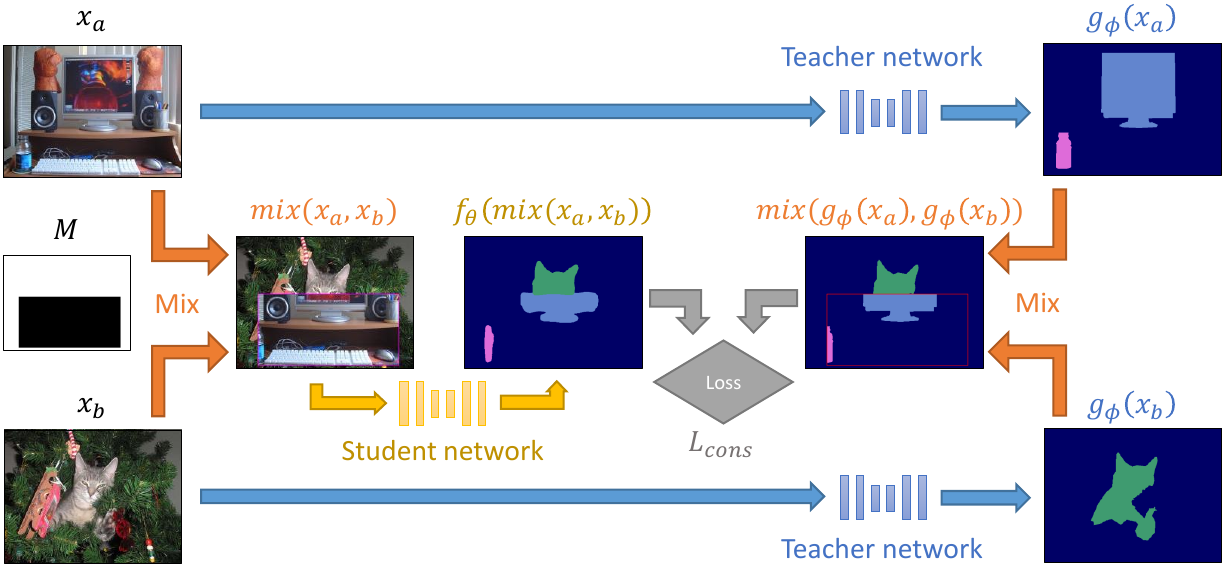}
\caption{\label{fig:supp:cutmix}%
Illustration of mixing regularization for semi-supervised semantic segmentation with the mean teacher framework.
$f_\theta$ and $g_\phi$ denote the student and teacher networks, respectively. The arbitrary
mask $M$ is omitted from the argument list of function $\mathit{mix}$ for legibility.
}
\end{figure}

\begin{figure}[t]
\centering
\includegraphics[width=.9\textwidth,trim={2mm 106mm 88mm 3.5mm},clip]{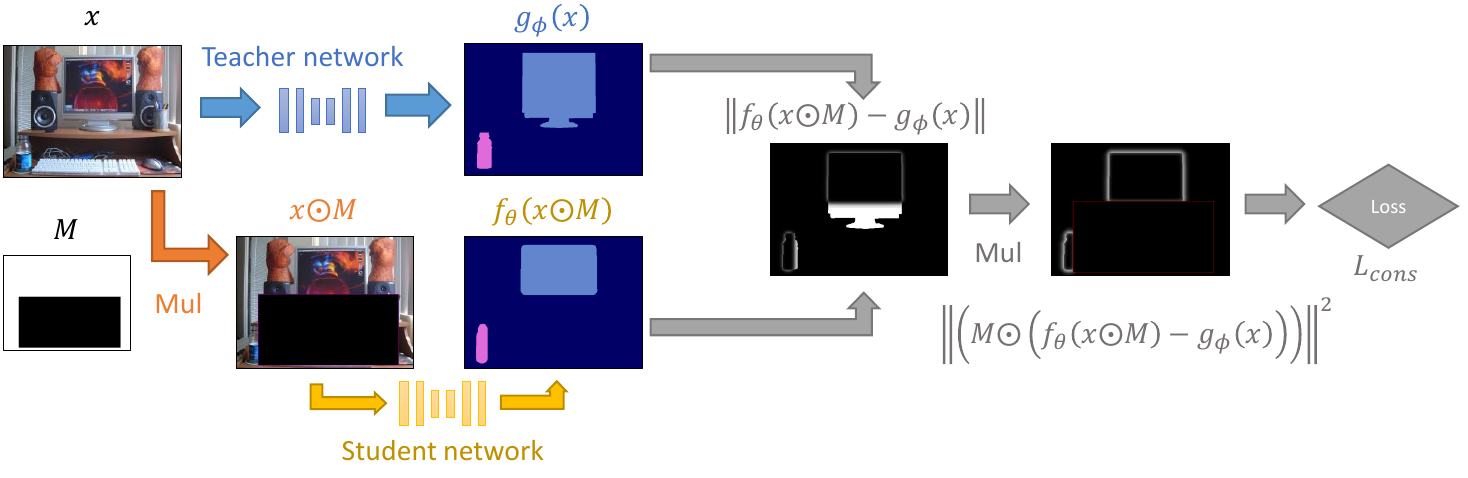}
\caption{\label{fig:supp:cutout}%
Illustration of Cutout regularization for semi-supervised semantic segmentation with the mean teacher framework.
Note that we include additional detail in final steps of the computation of $L_\textit{cons}$ in comparison to Figure~\ref{fig:supp:cutmix}
in order to illustrate the masking of the consistency loss.
}
\end{figure}

We illustrate the computation of CutMix based consistency loss $L_\textit{cons}$ in Figure~\ref{fig:supp:cutmix} and Cutout consistency loss in Figure~\ref{fig:supp:cutout}.

\subsection{CutMix with full-sized crops on \Cityscapes{}}

\begin{figure}[t]
\centering\footnotesize
\begin{tabular}{cc}
Image $A$  & Image $B$  \\
\includegraphics[width=0.45\textwidth]{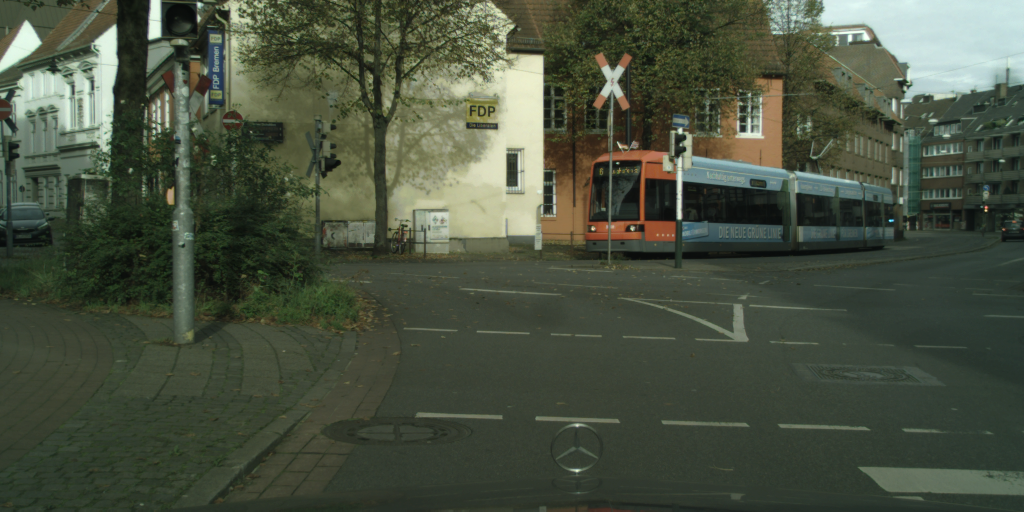}&%
\includegraphics[width=0.45\textwidth]{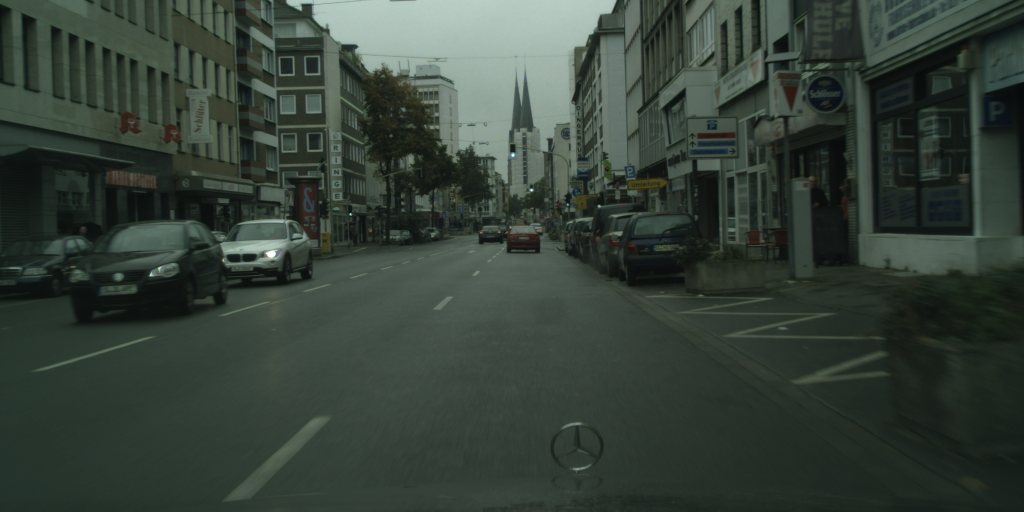}\\

Mask with one box  & Mask with three boxes combined using \texttt{xor}  \\
\fbox{\includegraphics[width=0.45\textwidth]{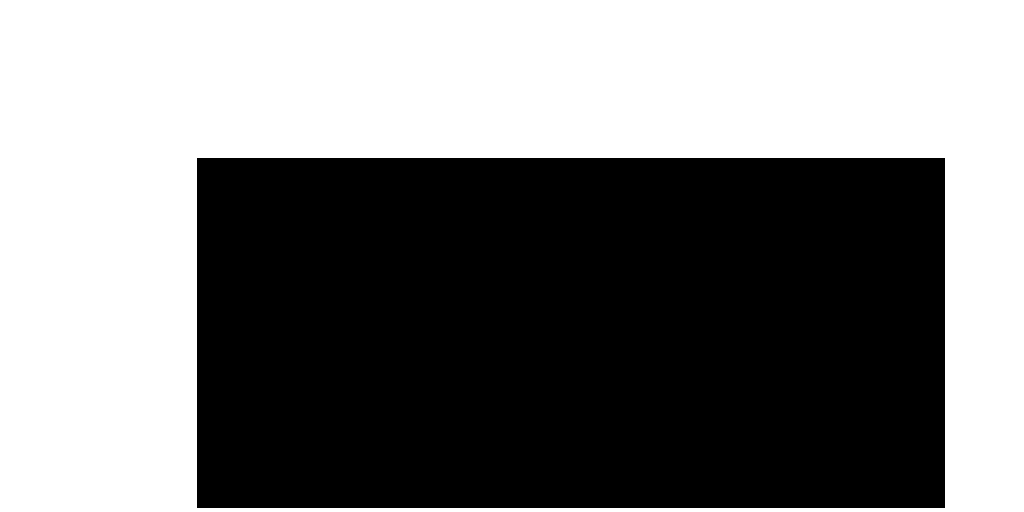}}&%
\fbox{\includegraphics[width=0.45\textwidth]{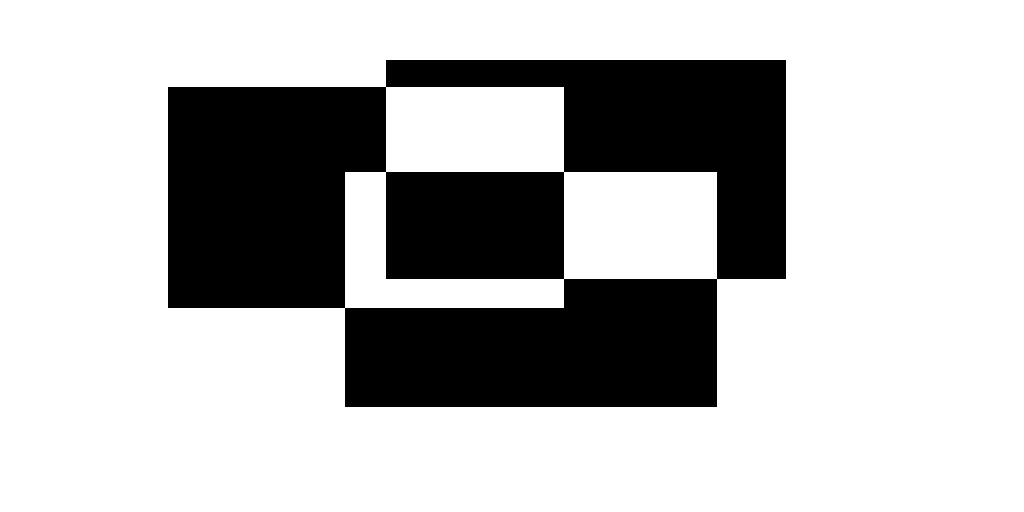}}\\

Mix of $A$ and $B$ using one box  & Mix of $A$ and $B$ using three boxes  \\
\includegraphics[width=0.45\textwidth]{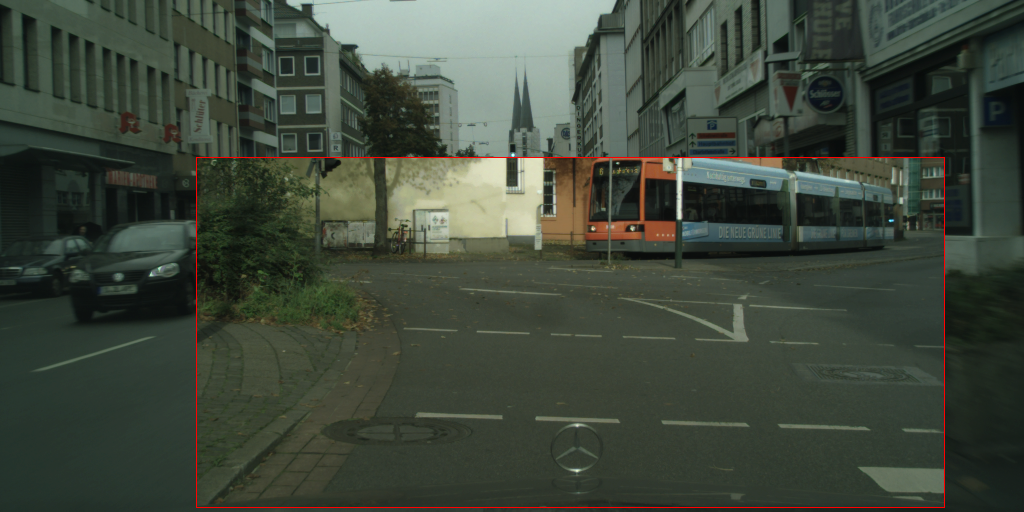}&%
\includegraphics[width=0.45\textwidth]{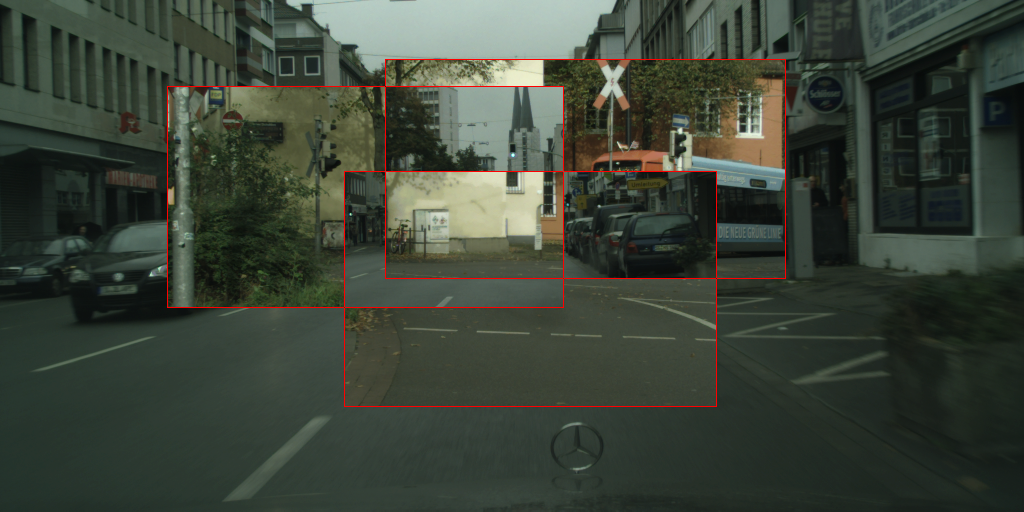}\\

\end{tabular}
\caption{\label{fig:supp:cutmix_threeboxes}%
CutMix using a one-box mask vs a three-box mask when using full image size crops from Cityscapes.
}
\end{figure}

As stated in our main text, when using the \Cityscapes{} dataset, using full size image crops -- $1024\times512$ rather than the usual $512\times256$ --
impairs the performance of semi-supervised learning using CutMix regularization, reducing the mIoU score from $60.34\%\pm1.24$ to $58.75\%\pm0.75$.
We believe that optimal performance is obtained when the scale of the elements in the mixing mask are appropriately matched to the scale of the image content.
We can alleviate this reduction in perofmrnace by constructing our mixing mask by randomly choosing three smaller boxes whose area is $\nicefrac{1}{3}$
of that used for one box (the normal case).
Given that a CutMix mask consisting of a single box uses a box that covers 50\% of the image area (but with random aspect ratio and position),
the three boxes each cover $\nicefrac{1}{6}$ of the image area.
The masks for the three boxes are combined using an \texttt{xor} operation.
Figure~\ref{fig:supp:cutmix_threeboxes} contrast mixing with one-box and three-box masks.

\subsection{Training details}
\label{app:seg:training}

\subsubsection{Using ImageNet pre-trained DeepLab v2 architecture for Cityscapes and Pascal VOC 2012}

We use the Adam \cite{Kingma:Adam} optimization algorithm with a learning rate of $3 \times 10^{-5}$. As per the mean teacher algorithm \cite{Tarvainen:MeanTeachers}, after each iteration the weights $w_t$ of the teacher network are updated to be the exponential moving average of the weights $w_s$ of the student: $w_t = \alpha_t w_t + (1 - \alpha_t) w_s$, where $\alpha_t=0.99$.

The \Cityscapes{} images were downsampled to half resolution ($1024 \times 512$) prior to use, as in \cite{Hung:AdvSemiSupSeg}. We extracted $512 \times 256$ random crops, applied random horizontal flipping and used a batch size of 4,
in keeping with \cite{Mittal:SSSHiLow}.

For the \Pascal{}~VOC experiments, we extracted $321 \times 321$ random crops, applied a random scale between 0.5 and 1.5 rounded to the nearest 0.1 and applyed random horzontal flipping. We used a batch size of 10, in keeping with \cite{Hung:AdvSemiSupSeg}.

We used a confidence threshold of 0.97 for all experiments. We used a consistency loss weight of 1 for both CutOut and CutMix, 0.003 for standard augmentation, 0.01 for ICT and 0.1 for VAT.

Hyper-parameter tuning was performed by evaluating performance on a hold-out validation set whose samples were drawn from the \Pascal{} training set.

We trained for 40,000 iterations for both datasets. We also found that identical hyper-parameters worked well for both using DeepLab v2.

\subsubsection{Using ImageNet pre-trained DenseUNet for ISIC 2017}

All images were scaled to $248\times248$ using area interpolation as a pre-process step. Our augmentation scheme consists of random $224\times224$ crops, flips, rotations and uniform scaling in the range 0.9 to 1.1.

In contrast to \cite{Li:SemiSupSkin} our standard augmentation based experiments allow the samples passing through the teacher and student paths to be arbitrarily rotated and scaled with respect to one another (within
the ranges specified above), where as \cite{Li:SemiSupSkin} use rotations of integer multiples of 90 degrees and flips.

All of our ISIC 2017 experiments use SGD with Nesterov momentum \cite{Sutskever:InitMomentum} (momentum value of 0.9) with a learning rate of 0.05 and weight decay of $5 \times 10^{-4}$.
For Cutout and CutMix we used a consistency weight of 1, for standard augmentation 0.1 and for VAT 0.1.

We would like to note that scaling the shortest dimension of each image to 248 pixels while preserving aspect ratio reduced performance;
the non-uniform scale in the pre-processing step acts as a form of data augmentation.

\subsubsection{Different architectures for augmented Pascal VOC 2012}

We found that different network architectures gave the best performance using different learning rates, presented in Table~\ref{tab:sup:lrs}.

\begin{table}[t]
\begin{center}%
\begin{tabular}{@{ }ll@{ }}
\hline
Architecture 							& Learning rate 				\\
\hline
DeepLab v2 								& $3 \times 10^{-5}$			\\
DeepLab v3+ 							& $1 \times 10^{-5}$			\\
DenseNet-161 based Dense U-net 			& $3 \times 10^{-4}$			\\
ResNet-101 based PSPNet 				& $1 \times 10^{-4}$			\\

\hline
\end{tabular}%
\caption{Learning rates used for different architectures, for the Pascal VOC 2012 dataset. All networks used pretrained weights for ImageNet classification.
}
\label{tab:sup:lrs}
\end{center}
\end{table}

We used the MIT CSAIL implementation\footnote{Available at \url{https://github.com/CSAILVision/semantic-segmentation-pytorch}.} of ResNet-101 based PSPNet~\cite{Zhao:PSPNet}.
We had to modify\footnote{Our modified version can be found in the \path{logits-from-models} branch of \url{https://github.com/Britefury/semantic-segmentation-pytorch}.} their code in order to use our loss functions. 
We note that we did \emph{not} use the \emph{auxiliary loss} from \cite{Zhao:PSPNet}, known as the \emph{deep supervision trick} in the MIT CSAIL GitHUb repository.

\subsubsection{Confidence thresholding}

\cite{French:SelfEnsDomAdapt} apply confidence thresholding, in which they mask consistency loss to 0 for samples whose confidence as predicted by the teacher network is below a threshold of 0.968. In the context of segmentation, we found that this masks pixels close to class boundaries as they usually have a low confidence. These regions are often large enough to encompass small objects, preventing learning and degrading performance. Instead we modulate the consistency loss with the proportion of pixels whose confidence is above the threshold. This values grows throughout training, taking the place of the sigmoidal ramp-up used in \cite{Laine:Temporal, Tarvainen:MeanTeachers}.

\subsubsection{Consistency loss with squared error}

Most implementations of consistency loss that use squared error (\eg \cite{Tarvainen:MeanTeachers}) compute the mean of the squared error over all dimensions.
In contrast we sum over the class probability dimension and computing the mean over the spatial and batch dimensions.
This is more in keeping with the definition of other loss functions use with probability vectors such as cross-entropy and KL-divergence.
We also found that this reduces the necessity of scaling the consistency weight with the number of classes; as is required then taking the mean over the class probability dimension \cite{Tarvainen:MeanTeachers}.

\end{document}